\documentclass[conference]{IEEEtran}
\IEEEoverridecommandlockouts

\usepackage{amsmath,amssymb,amsfonts}
\usepackage{algorithmic}
\usepackage{graphicx}
\usepackage{textcomp}
\usepackage{xcolor}
\usepackage[noadjust]{cite}

\usepackage{graphicx}
\usepackage{multirow}
\usepackage{booktabs}
\usepackage{adjustbox}

\definecolor{bleudefrance}{rgb}{0.19, 0.55, 0.91}

\newcommand{\real}{\mathbb{R}}
\newcommand{\mat}[1]{\boldsymbol{#1}}
\newcommand{\norm}[1]{\left\lVert #1 \right\rVert}

\def\BibTeX{{\rm B\kern-.05em{\sc i\kern-.025em b}\kern-.08em
    T\kern-.1667em\lower.7ex\hbox{E}\kern-.125emX}}
\begin{document}

\title{Guided Model-based LiDAR Super-Resolution for Resource-Efficient Automotive scene Segmentation
\thanks{This work has received funding from the EU’s Horizon Europe research and innovation programme in the frame of the AutoTRUST project “Autonomous self-adaptive services for TRansformational personalized inclUsivenesS and resilience in mobility” under the Grant Agreement No 101148123.}
}

\author{
\IEEEauthorblockN{Nikos Piperigkos$^{1,2}$, Alexandros Gkillas$^{1,2}$, Christos Anagnostopoulos$^{1,2}$, Aris S. Lalos$^{1}$}
\IEEEauthorblockA{$^1$Industrial Systems Institute, Athena Research Center, Patras Science Park, Greece\\
$^2$AviSense.AI, Patras Science Park, Greece\\
Emails: \{piperigkos, gkillas, anagnostopoulos\}@avisense.ai, lalos@isi.gr
}
}

\author{
\IEEEauthorblockN{Alexandros Gkillas$^{1,2}$, Nikos Piperigkos$^{1,2}$, Aris S. Lalos$^{1}$}
\IEEEauthorblockA{$^1$Industrial Systems Institute, Athena Research Center, Patras Science Park, Greece\\
 $^2$AviSense.AI, Patras Science Park, Greece\\
 Emails:\{ gkillas, piperigkos\}@avisense.ai, lalos@isi.gr
 }
}

\maketitle

\begin{abstract}
High-resolution LiDAR data is essential for 3D semantic segmentation in autonomous driving. However, the high cost of advanced LiDAR sensors limits their widespread adoption. Affordable alternatives like 16-channel LiDAR generate sparse point clouds, resulting in reduced segmentation performance. To address this, we present the first end-to-end framework that integrates LiDAR super-resolution (SR) and segmentation tasks. This framework achieves context awareness through a novel joint optimization during training, ensuring that the SR process integrates semantic guidance from the segmentation task and preserves the details of smaller classes. Furthermore, the proposed framework incorporates a novel SR loss function to enhance the network’s focus on regions of interest. The lightweight model-based SR network achieves a significant reduction in number of parameters compared to state-of-the-art LiDAR SR models, enabling efficient integration with segmentation networks. Experimental results demonstrate that our method delivers segmentation performance comparable to networks using high-resolution and high-cost 64-channel LiDAR data.
\end{abstract}

\begin{IEEEkeywords}
LiDAR, 3D segmentation. super-resolution, model-based deep learning, range view
\end{IEEEkeywords}

\section{Introduction}
\label{sec:intro}

LiDAR technology is a foundational element of autonomous vehicles, enabling reliable operation in challenging environmental conditions. By capturing detailed 3D point clouds, LiDAR provides data necessary for perception systems to  understand their surroundings, ensuring safe  navigation in complex environments \cite{9127855}. To this end, 
semantic segmentation plays a crucial role  by assigning labels to individual points within a point cloud, thereby transforming raw data into a structured representation of the environment \cite{jhaldiyal2023semantic}.  However, the accuracy  of the segmentation results are heavily influenced by the resolution and quality of the LiDAR data, highlighting the importance of high-resolution 3D input for efficient segmentation \cite{jhaldiyal2023semantic}. 
In particular, providing high-resolution LiDAR data is challenging due to the prohibitive cost of advanced sensors \cite{SHAN2020103647}. 
For example, a 64-channel HDL-64E LiDAR costs around $85,000\$$, while 16-channel sensors are more affordable at $4,000\$$ but produce lower-resolution data. 

Although recent works have increasingly focused on enhancing the resolution of low-cost LiDAR sensors to achieve performance comparable to high-resolution counterparts, existing super-resolution (SR) approaches do not explore their impact on perception tasks, such as segmentation. 
These methods typically operate either directly on 3D point cloud data or on 2D range images derived from LiDAR scans \cite{9875347, yang2024tulip}. However, despite their potential, these approaches face several significant limitations. First, existing SR methods rely heavily on deep learning models with complex architectures and numerous learnable parameters, requiring large amounts of high-quality training data \cite{9284628, 9173575}. Additionally, these methods operate independently of perception tasks e.g., the segmentation, meaning that the training of SR networks is not influenced by the segmentation process. This independent optimization approach creates additional challenges when the SR network's output is used as input to a separately trained segmentation network. As a result, the SR methods often produce outputs containing high level of outliers, which heavily degrade  segmentation performance \cite{SHAN2020103647}. Additionally, the aforementioned methods fail to preserve details about smaller classes, since the SR process inherently put more emphasis on dominant categories (e.g., buildings). Consequently, the sparse representations of minority classes become distorted, degrading segmentation performance.
Hence, optimizing the SR and segmentation networks separately leads to a significant decline in segmentation accuracy, demonstrating the necessity of a joint optimization approach within an end-to-end framework.

Nevertheless, to achieve a resource-efficient end-to-end architecture, the SR model should be as lightweight as possible to avoid adding computational complexity to the segmentation process. Unlike existing methods that employ complex deep learning models  (e.g., \cite{yang2024tulip}), our approach draws its motivation from the model-based deep learning theory \cite{model_eldar}, leading a more efficient solution for the SR problem. To be more detailed, based on a novel optimization problem, we design a resource efficient model-based SR network utilizing the deep unrolling strategy \cite{model_eldar}. The resulting SR model can be integrated with the segmentation network within a unified system. By jointly optimizing the SR and segmentation networks during the training phase, the SR model-based network effectively preserves critical contextual details required for accurate segmentation by incorporating semantic information  provided by the segmentation network. To the best of our knowledge, this is the first study to propose such an end-to-end framework.
Therefore, the contributions of this work are 
\begin{itemize} 


\item \textbf{Guided Model-based SR Network:} We introduce a novel optimization problem for the LiDAR super-resolution task, formulated in the range-view domain. The cost function of the optimization framework consists of three key components: (1) a data-consistency module that enforces the  relationship between the low-resolution and high-resolution range images, (2) an encoder-decoder-based regularizer that captures intrinsic properties of high-resolution range images, such as low-rank structures and data similarities, and (3) a learnable mask module guided by segmentation information during training. This learnable mask integrates  segmentation guidance into the SR process and is adapted for inference to provide context-awareness. The iterative solution of the proposed optimization problem is transformed into a model-based deep learning network using the deep unrolling strategy, resulting in a computationally efficient network.

\item \textbf{End-to-End architecture}: The proposed  SR model is combined with the segmentation network to form the end-to-end framework. Through the joint optimization of both the SR and segmentation networks, the SR module effectively preserves key regions of the scene  for accurate segmentation. Overall, the joint optimization and lightweight nature of the SR model allow the resulting end-to-end framework to achieve performance comparable to segmentation networks that utilize high-resolution and high-cost 64-channel LiDAR sensors.

\item \textbf{Context-Aware Loss Function}: To enhance further the ability of the proposed SR network to preserve the structure of underrepresented classes, we introduce a context-aware loss function during the end-to-end training process. This loss function leverages segmentation masks generated by the segmentation network to guide the SR model to focus more on key areas of interest.
\end{itemize}

\section{Related Work}
This section provides a brief overview of relevant research on 3D LiDAR segmentation and  LiDAR SR methods. 
\subsection{LiDAR-based Segmentation}
Segmentation methods have been developed, typically categorized into point-based, voxel-based, and projection-based approaches.
\textbf{Point-based} methods process raw 3D points directly, preserving spatial structure without transformation of the points utilizing deformable convolutions with an arbitrary number of kernel points to capture local geometric structures\cite{Hu_2020_CVPR}. 
\textbf{Voxel-based} methods tackle the irregularity of 3D point clouds by discretizing the space into uniform voxel grids, allowing for the use of convolutional operations to predict semantic labels. such as  \cite{cheng2021af2s3netattentivefeaturefusion}. Despite these improvements, voxel-based and point-based methods still suffer from the sparsity of point clouds, leading to redundant calculations and high memory consumption.
\textbf{Projection-based methods}  transform 3D point clouds into 2D image representations, enabling the use of advanced image feature extraction techniques for semantic segmentation \cite{ Kong_2023_ICCV}.
Methods such as FIDNet \cite{zhao2021fidnetlidarpointcloud} and CENet \cite{9859693} enhance segmentation performance by using ResNet-based encoders and simplifying decoders with interpolation techniques. Similarly, LENet \cite{lenet}, which is a lighweight version of CENet \cite{9859693}, achieves state-of-the-art results by integrating the MSCA (Multi-Scale Context Aggregation) and IAC (Inter-Scale Attention Calibration) modules, which  boost the overall performance.

\subsection{LiDAR-based Super-resolution}
LiDAR super-resolution techniques can be broadly divided into two main approaches. The first category focuses on performing super-resolution directly on raw 3D LiDAR point clouds \cite{9919373, 9875347}. Nevertheless, these methods generally require substantial computational resources \cite{ZHENG2024103783}.
Alternative approaches \cite{9811992, 10164213, SHAN2020103647} focus on the range image domain by projecting 3D point clouds onto 2D range images. However, these methods typically depend on deep learning models with a large number of parameters such as U-Net-based networks \cite{SHAN2020103647} and swin transformers \cite{yang2024tulip}. 
Additionally, the high computational complexity of existing SR methods, coupled with the need for post-processing to eliminate outliers, hinders their ability to achieve a satisfactory frame rate (fps). 
Furthermore, the existing approaches struggle to maintain the structural details of smaller  classes, when the SR and segmentation networks are optimized independently, limiting their ability to accurately reconstruct key regions. An end-to-end framework offers a viable solution to this problem. However, the computational demands of current SR methods restrict their feasibility for such frameworks, underscoring the need for an efficient SR model that achieves high performance without excessive computational overhead.

To this end, this work tackles the computational complexity of existing SR methods by introducing a novel optimization problem tailored for the super-resolution task in the range-view domain. The proposed optimization framework is efficiently transformed into a model-based SR network through the deep unrolling strategy.  While similar to the approach in \cite{10222856}, their method was agnostic to the segmentation task. As a result, it faced challenges such as producing outliers and failing to preserve the structural integrity of smaller or underrepresented classes.
In our work, the derived SR model is guided by the segmentation network. This lightweight super-resolution model is jointly trained with the segmentation network, forming the proposed end-to-end framework. The joint optimization process ensures that the SR network benefits from segmentation guidance, preserving structural consistency throughout the SR process and significantly enhancing segmentation accuracy.

\section{Backbone Segmentation Architecture} \label{segmentation}


\begin{figure*}
\centering
 \includegraphics[scale=0.4]{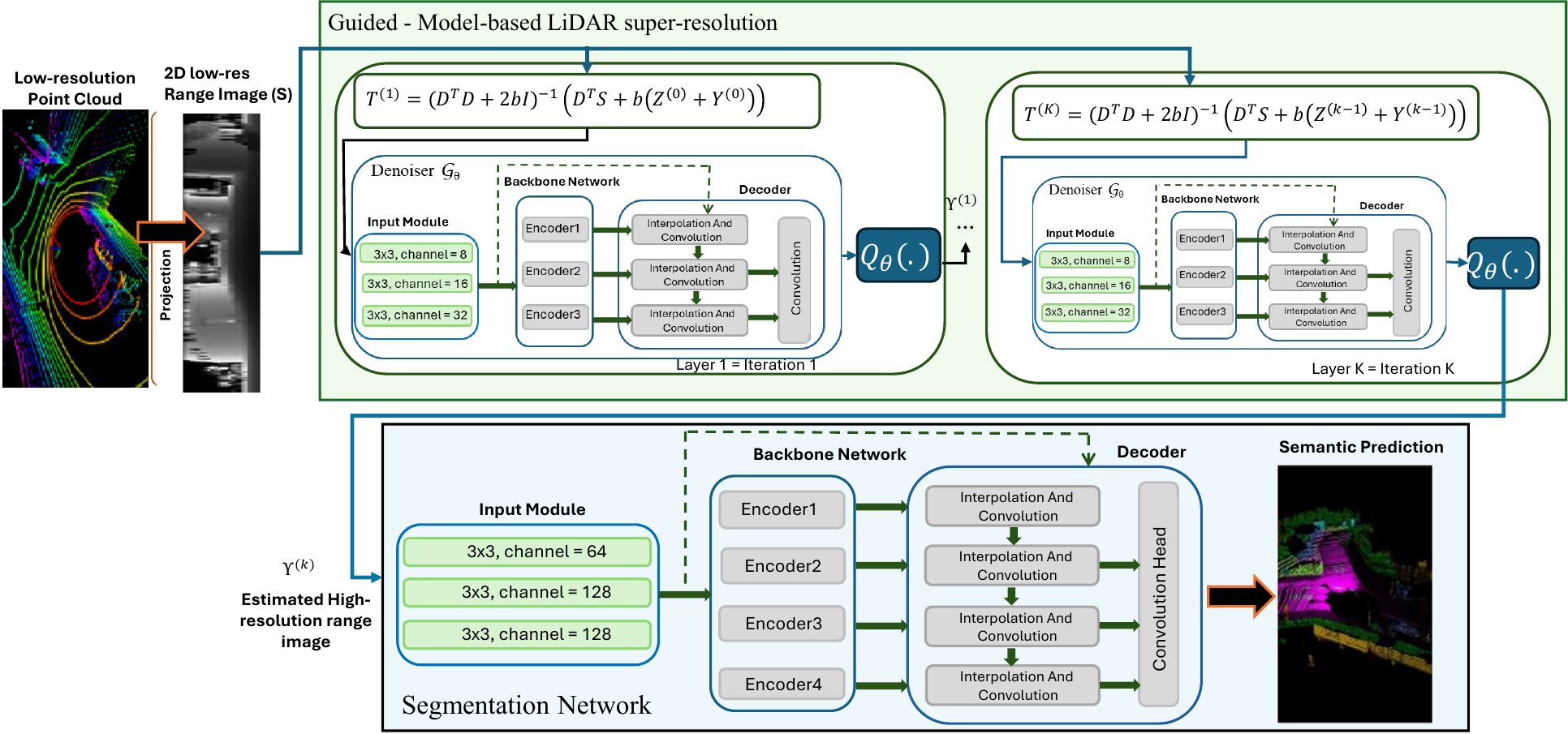}
  \caption{Illustration of the key components of the proposed end-to-end architecture. (a) \textbf{Segmentation network}:The segmentation network employs a hierarchical backbone structure inspired by ResNet34. Its architecture incorporates the IAC module, which progressively upsamples low-resolution feature maps to their original dimensions while integrating outputs from preceding IAC modules.
  (b) \textbf{Guided Model-based SR network:} A small number of iterations of the solver in (\ref{eq:HQS_final}) are unrolled
and treated as a deep learning architecture, consisting of the data-consistency solution (\ref{eq:x_up}), the denoiser (\ref{eq:z_up}) and and the segmentation-guided regularization using the learnable mask (\ref{eq:y_up}).
(c) Overall \textbf{end-to-end architecture:} A 16-channel LiDAR point cloud is converted into a low-resolution range image, processed by the SR network to generate a high-resolution range image. This high-resolution output is fed into the segmentation network for final 3D segmentation. By jointly optimizing the SR and segmentation networks, the SR process benefits from  critical semantic guidance, thus  enhancing the segmentation performance.}
  \label{fig:LENet}
\end{figure*}

Considering computational efficiency and state-of-the-art performance in the 3D segmentation problem, we will focus on projection-based methods operating in the 2D range image domain. In light of this fact, we adopt LENet \cite{lenet}, a lightweght version of the CENet \cite{9859693}, as the backbone segmentation network for the proposed end-to-end architecture. In more detail, LENet integrates a Multi-Scale Convolution Attention (MSCA) module within its encoder and a lightweight Interpolation And Convolution (IAC) decoder for efficient processing of LiDAR data. The MSCA module consists of three main components: a depth-wise convolution for local information aggregation, multi-branch depth-wise strip convolutions for capturing multi-scale context, and a $1\times1$ convolution to model inter-channel relationships. The output of the $1\times1$  convolution acts as attention weights, refining the input features for enhanced representation. The encoder building block consists of a $3 \times 3$ convolution layer followed by the MSCA module. For decoding, LENet incorporates an IAC module, which employs bilinear interpolation for upsampling feature maps from the encoder, a $3\times3$  convolution to merge information from the encoder and previous IAC stages, and a point-wise convolution to fuse the outputs of the last three IAC modules. The complete encoder-decoder architecture is illustrated in Figure \ref{fig:LENet}.

\section{Proposed Methodology}
In this section, we present the proposed end-to-end architecture.
\subsection{Guided Model-based SR}

\subsubsection{Proposed optimization problem}
To construct the proposed model-based deep learning architecture, we leverage the advantages of projection-based methodologies. Specifically, we project the 3D high-resolution point cloud $\mat{P}$ obtained from a 64-channel LiDAR sensor, into a high-resolution range image $\mat{T} \in \real^{64\times N}$. In view of this, the  high-resolution data and the low-resolution data sensor are connected as follows:
\begin{equation} \mat{S} = \mat{D} \mat{T} + \mat{E}, \label{eq:degr_model} \end{equation}
where  $\mat{D} \in \real^{16 \times 64}$ is the downsampling operator that selects  the $16$ channels from the high resolution range image and $\mat{E}$ is a term.   A simple optimization to enhance the resolution of the range image is 
\begin{equation}
\underset{\mat{T}}{\arg\min}\,\,\, \frac{1}{2} \norm{\mat{S} - \mat{D} \mat{T}}_F^2  + \mu \mathcal{J}(\mat{T}), 
\label{eq:main_problem_seg}
\end{equation}
where the first component ensures consistency with the degradation model defined in \eqref{eq:degr_model}. The second component $\mathcal{J}(\cdot)$ serves as a learnable regularizer, designed to capture the intrinsic features of the high-resolution range image  $\mat{T}$.

However,  the above formulation remains agnostic to segmentation problem, resulting in a lack of influence from the segmentation results during the training and operational phase. This limitation can lead to the loss of structural details for smaller or underrepresented classes. To address this limitation, we introduce an additional regularization term in the optimization problem that leverages a learnable mask to guide the super-resolution process. This learnable mask is trained using the ground truth segmentation masks during training, ensuring a focus on regions associated with underrepresented classes. The  segmentation-guided optimization problem is defined as:
\begin{equation}
\underset{\mat{T}}{\arg\min}\,\,\, \frac{1}{2} \norm{\mat{S} - \mat{D} \mat{T}}_F^2  + \mu\mathcal{J}(\mat{T}) + \lambda \mathcal{Q}(\mat{T}),
\label{eq:main_problem_seg}
\end{equation}
where $\mathcal{Q}(\cdot)$
 represents an extra regularizer that guides the estimated high-resolution output to preserve the structure of the classes of interest for the segmentation task. 

\subsection{Guided Model-based deep learning SR} \label{DU}
To tackle the proposed optimization problem in (\ref{eq:main_problem_seg}), we employ the Half quadratic splitting (HQS) methodology. The problem can be reformulated as follows:
\begin{align}
&\underset{\mat{T}}{\arg\min}\,\,\, \frac{1}{2} \norm{\mat{S} - \mat{D} \mat{T}}_F^2  + \mu\mathcal{J}(\mat{Z}) + \lambda \mathcal{Q}(\mat{Y}) \nonumber \\
&s.t. \,\,\,\, \mat{T} = \mat{Z}, \mat{T} = \mat{Y},
\label{eq:main_problem_seg_hqs}
\end{align}
where $\mat{Z} \in \real^{64\times N}$, $\mat{Y}\in \real^{64\times N}$ are auxiliary variables. The loss function that HQS aims to minimize is:
\begin{align}
    \mathcal{L} = \frac{1}{2}\norm{\mat{S} - \mat{D} \mat{T}}_F^2  &+ \mu\mathcal{J}(\mat{Z}) + \lambda \mathcal{Q}(\mat{Y}) \nonumber\\
    &+ \frac{b}{2}\norm{Z-T}_F^2+ \frac{b}{2}\norm{Y-T}_F^2,
    \label{eq:Lagrangian}
\end{align}
where b denotes a penalty parameter. Based on equation (\ref{eq:Lagrangian}) a sequence of individual sub-problems emerges, that are analyzed in the following:
\begin{subequations}
\begin{align}
    \mat{T}^{(k+1)} = \underset{\mat{T}}{\arg\min}\,&\frac{1}{2} \norm{\mat{S} - \mat{D} \mat{T}^{(k)}}_F^2 
    + \frac{b}{2}\norm{\mat{Z}^{(k)} -\mat{T}^{(k)}}_F^2 \nonumber \\
    &+ \frac{b}{2}\norm{\mat{Y}^{(k)} -\mat{T}^{(k)}}_F^2 
     \label{eq:updateX},\\
   \mat{Z}^{(k+1)} = \underset{\mat{Z}}{\arg\min}\, &\mu \mathcal{J}(\mat{Z}) 
   + \frac{b}{2}\norm{\mat{Z} -\mat{T}^{(k+1)}}_F^2\ ,\label{eq:updateZ}\\
\mat{Y}^{(k+1)} =  \underset{\mat{Y}}{\arg\min}\,&\lambda \mathcal{Q}(\mat{Y}) 
   + \frac{b}{2}\norm{\mat{Y} -\mat{T}^{(k+1)}}_F^2\ \label{eq:updateY}
\end{align}
\end{subequations}

\textbf{Data consistency Module - subproblem (\ref{eq:updateX}):} The closed form solution is given as:
\begin{align}
    \mat{T}^{(k+1)} = (\mat{D}^T \mat{D} + 2b\mat{I})^{-1}(\mat{D}^T\mat{Y} + b\mat{Z}^{(k)} + b\mat{Y}^{(k)})
    \label{eq:x_solution}
\end{align}
\textbf{Denoising Module - subproblem (\ref{eq:updateZ}):} The purpose of the denoising module is to produce a refined range image by taking as input the estimated range image from the data consistency module in equation (\ref{eq:x_solution}). Thereby,  this step  can be implemented using a deep learning-based denoiser as follows:
\begin{equation} \mat{Z}^{(k+1)} = \mathcal{G}_\theta(\mat{T}^{(k+1)}), \label{eq:denoiser_z} 
\end{equation}
where $\mathcal{G}_\theta$ denotes the neural network responsible for denoising. 
For the architecture of the deep learning-based denoiser, we draw inspiration from the LENet segmentation network, as illustrated in Figure 1.

\textbf{Segmentation Guidance - subproblem (\ref{eq:updateY}):}
\textit{Although this subproblem originally involves the variables \(\mat{Y}\) and \(\mat{T}\), we substitute \(\mat{T}\) with \(\mat{Z}\), which represents the denoised version of \(\mat{T}\) obtained from the previous step. This substitution provides a cleaner and more precise representation.} The solution of this problem can be replaced again with a lernable function. 
To this end, we introduce a learnable mask \(\mathcal{Q}_\theta(\cdot)\) to incorporate segmentation guidance across both the training and inference phases, based on the solution $\mat{Z}^{(K+1)}$ from step (8).  
During training, the segmentation mask derived from the ground truth (\(M_{\text{GT}}\)) supervises the learning of \(\mathcal{Q}_\theta(.)\), ensuring accurate segmentation guidance. The loss for the learnable mask is 
\begin{equation}
    \mathcal{L}_{\text{mask}} = \frac{1}{N} \sum_{j=1}^N \norm{ \mathcal{Q}_\theta(Z^{(k+1)}(j)) - w_{c(j)} M_{\text{GT}}(j) }_1,
    \label{eq:segloss}
\end{equation}
where \(N\) is the number of pixels, \(Z^{(k+1)}\) is the input  at iteration \(k+1\), and \(M_{\text{GT}}\) is the ground truth segmentation mask. The \(c(j)\) is  class label of pixel \(j\) , as indicated by the segmentation mask \(M_{GT}\). The weight $w_{c(j)} = \frac{1}{\sqrt{f_{c(j)}}}$, derived from the class frequency $f_{c(j)}$, ensures higher emphasis on underrepresented classes.
 During inference, where \(M_{\text{GT}}\) is unavailable, the learned neural network \(\mathcal{Q}_\theta(.)\) generates the mask based on the input \(Z^{(k+1)}\) with dimension $64 \times 1024$. This estimated mask is then used to enhance context-awareness by applying a pixel-wise multiplication with \(\mat{Z}^{(k+1)}\), ensuring that segmentation guidance is explicitly incorporated.
The updated subproblem is formulated as:
\begin{equation}
    \mat{Y}^{(k+1)} = \mathcal{Q}_\theta(\mat{Z}^{(k+1)}) \odot \mat{Z}^{(k+1)}
\end{equation}
\textbf{Final Iterative solutions}:
Hence, the HQS solver  consists of three modules that is  the data consistency  solution for estimating the high-resolution range image (\ref{eq:x_up}), the denoising step in equation (\ref{eq:z_up}) and the learnable segmentation mask that guides the output to preserve critical  structure details (\ref{eq:y_up}).
\begin{subequations}
\begin{align}
\mat{{T}}^{(k+1)}=(\mat{D}^T \mat{D} + b\mat{I})^{-1}(\mat{D}^T\mat{\mat{S}} + b(\mat{{Z}}^{(k)} +\mat{{Y}}^{k})) \label{eq:x_up}\\
\mat{Z}^{(k+1)}=\mathcal{G}_\theta(\mat{T}^{(k+1)}) \label{eq:z_up}\\
\mat{Y}^{(k+1)}=\mathcal{Q}_\theta(\mat{Z}^{(k+1)}) \odot \mat{Z}^{(k+1)}\label{eq:y_up}
\end{align}
\label{eq:HQS_final}
\end{subequations} 
\textbf{Guided-Model-based SR network}:
Since the denoiser contain learnable parameters, the solutions of the proposed optimization problem can be reformulated into a computationally efficient architecture. To achieve this, we adopt the Deep Unrolling (DU) framework. From equation (\ref{eq:HQS_final}), we unroll only K iterations and treat each one as a single layer in the resulting deep learning model. The proposed model-based SR network is illustrated  in Figure \ref{fig:LENet} along with the overall end-to-end architecture that we analyze in the next Section.

\subsection{End-to-End Architecture}

Having designed the proposed model-based SR network with low computational complexity, we now introduce the complete end-to-end architecture, as illustrated in Figure \ref{fig:LENet}. The process begins with a low-resolution point cloud captured by a 16-channel LiDAR sensor, which is converted into a low-resolution range image. This range image is then fed into the model-based SR network, described in Section \ref{DU}, to generate a high-resolution range image. The high-resolution range image is subsequently passed to the segmentation network described in Section \ref{segmentation} for final 3D semantic segmentation. By jointly optimizing the SR and segmentation networks, the end-to-end architecture incorporates semantic context into the SR process, enhancing the quality of the super-resolved data. 

\subsubsection{End-to-End Optimization}
A key element of the proposed architecture is the end-to-end training of the SR  and segmentation networks. \textbf{Context-Aware LiDAR Loss:}
To enhance the ability of the SR network to preserve the structure of smaller classes, we introduce a context-aware loss function. This loss function leverages the ground truth segmentation masks to guide the SR model’s focus, defined as
\begin{align}
L_{sr} = \sum_{i=1}^{p} \sum_{j=1}^{N} w_{c(j)} \norm{Y_i^{(K+1)}(j) - T_i(j) }_1
\end{align}
where \(p\) denotes the  number of range images and \(N\) the number of pixels per image. The weight $w_{c(j)}$ is defined similar to equation (\ref{eq:segloss}). \(Y_i^{(K+1)}(j)\) is the predicted  pixel \(j\) in the \(i\)-th image, while \(T_i(j)\) is the real value.
\textbf{Segmentation Loss:}
A  cross-entropy loss $L_{\text{wce}}$ is employed defined as
\begin{equation}
    L_{\text{wce}}(y, \hat{y}) = - \sum_{i} \alpha_i \cdot p(y_i) \log(p(\hat{y}_i))
\end{equation}
where $y_i$ and $\hat{y}_i$ represent the ground truth and predicted class labels, respectively.
The total loss is given by
\begin{equation}
    L = w_1 L_{\text{wce}} + w_2 L_{\text{ls}} + w_3 L_{sr} + w_4 L_{mask},
\end{equation}
where $L_{mask}$ is defined in equation (\ref{eq:segloss}), the  $w_1$, $w_2$, and $w_3$ are weights set empirically to $w_1 = 1$, $w_2 = 1.5$, and $w_3, w_4 = 1$. 


\section{Numerical results}
\subsection{Simulation setup}
\subsubsection{Dataset}
We utilize two widely-used LiDAR segmentation benchmarks: the \textbf{SemanticKITTI} dataset~\cite{9010727} and the \textbf{SemanticPOSS} dataset~\cite{10.1109/IV47402.2020.9304596}. SemanticKITTI is a large-scale dataset collected using the Velodyne HDL-64E LiDAR sensor, specifically designed for point cloud segmentation in autonomous driving scenarios. It is split into three subsets: sequences 00--04 for training, sequence 05 for validation, and sequences 06--08 for testing. SemanticPOSS, uses a PANDORA 40 channel LiDAR sensor.
For SemanticKITTI, we project point clouds from the 64-channel sensor into high-resolution range images of size $64 \times 1024$, which serve as ground truth. Corresponding low-resolution range images of size $16 \times 1024$ are generated by selecting 16 out of the 64 channels, simulating data from a low-cost 16-channel LiDAR. A similar process is applied to the SemanticPOSS dataset, producing high-resolution range images of size $40 \times 1024$ and low-resolution versions of size $10 \times 1024$.

\begin{table*}
  \caption{PERFORMANCE COMPARISON ON SEMANTICKITTI  BENCHMARK}
  \resizebox{\linewidth}{!}{
  \begin{tabular}{cccccccccccccccccl}
    \toprule
    & SR  & Segmentation &  FPS & Car & Bicycle & Motorcycle & Truck & Person  & Road  & Parking   & Building & Fence  & Trunk & Terrain & Traffic-Sign \\
    & Params (M) & Params (M) &  & &  & &  &   &   &    &  &  &  &  &  & &  \\
        \midrule
      LiDAR 64-channel & - & 4.7 & 27 & 0.97 & 0.38 & 0.72  & 0.61 & 0.47 & 0.94 & 0.62  & 0.89  & 0.63   & 0.64 & 0.65  & 0.48 \\
    \midrule
      LiDAR 16-channel &- &4.7 & 29 & 0.81 & 0.13 & 0.35 & 0.37 & 0.12 & 0.84 & 0.54  & 0.73  & 0.47    & 0.48 & 0.54  & 0.22 \\
    \midrule
      Transformer \cite{yang2024tulip} + LeNET &50  & 4.7& 6 &0.88 & 0.07 & 0.40 & 0.39 & 0.16 & 0.89 & 0.53 & 0.81  & 0.49    & 0.50 & 0.58  & 0.24 \\
    \midrule
      Proposed End-to-End & \textbf{0.1}& 4.7 & 23 &\textbf{0.90} & \textbf{0.37} & \textbf{0.60} & \textbf{0.60} & \textbf{0.44} & \textbf{0.93} & \textbf{0.59}  & \textbf{0.83}  & \textbf{0.58}   & \textbf{0.56} & \textbf{0.63}  & \textbf{0.45} \\
    \bottomrule
  \end{tabular}}
    \label{tab:results_deep}
\end{table*}
\subsubsection{Implementations Details}
For the proposed model-based SR network, we unroll the derived HQS solver in equation (\ref{eq:HQS_final}) for \(k=4\) iterations, resulting in a 4-layer deep learning architecture. 
 In the end-to-end framework, we use the AdamW optimizer with default PyTorch settings. The initial learning rate is set to \(2 \times 10^{-3}\) and dynamically adjusted using a cosine annealing scheduler over 80 epochs.
Also,  we utilize the mean Intersection over Union (mIoU) metric to evaluate the performance \cite{zhao2021fidnetlidarpointcloud}. Finally, all experiments were conducted on an NVIDIA RTX 4090 GPU.
\subsection{Evaluation study}

Table~\ref{tab:results_deep} and ~\ref{tab:results_deep_pos} show the quantitative results on the SemanticKITTI and  SemanticPOSS benchmarks for the LeNET segmentation network across different configurations: using the high-resolution LiDAR sensor, the low-resolution  LiDAR sensor, the state-of-the-art (SOTA) Transformer-based LiDAR SR method \cite{yang2024tulip} combined with the LeNET segmentation network, and the proposed end-to-end architecture.

\begin{table}
  \caption{Performance comparison on SemanticPOSS benchmark}
  \label{tab:results_deep_pos}
  \centering
  \resizebox{\linewidth}{!}{
  \begin{tabular}{lccccccccccc}
    \toprule
    Method & Person & Rider & Car & Truck & Plants & Traffic sign & Pole & Building & Bike & Ground \\
    \midrule
    LiDAR 40-channel & 0.76 & 0.24 & 0.77 & 0.67 & 0.74 & 0.54 & 0.32 & 0.81 & 0.53 & 0.80 \\
    LiDAR 10-channel & 0.52 & 0.05 & 0.43 & 0.25 & 0.66 & 0.15 & 0.28 & 0.72 & 0.34 & 0.70 \\
    Transformer \cite{yang2024tulip} + LeNET & 0.61 & 0.07 & 0.56 & 0.26 & 0.68 & 0.08 & 0.29 & 0.73 & 0.40 & 0.72 \\
    Proposed & \textbf{0.69} & \textbf{0.16} & \textbf{0.75} & \textbf{0.63} & \textbf{0.72} & \textbf{0.53} & \textbf{0.31} & \textbf{0.76} & \textbf{0.43} & \textbf{0.75} \\
    \bottomrule
  \end{tabular}}
\end{table}

\begin{table}
\centering
\caption{Impact of the Proposed Context-Aware SR Loss on Segmentation Performance -  SEMANTICKiTTI}
\label{tab:context_aware_loss}
\resizebox{\linewidth}{!}{%
\begin{tabular}{lccc}
\toprule
\textbf{Class} & \textbf{Without Context-Aware SR Loss} & \textbf{With Context-Aware SR Loss} & \textbf{Improvement (\%)} \\
\midrule
Person         & 0.36                                   & 0.44                                & +22.22 \\
Bicycle        & 0.30                                   & 0.37                                & +23.33 \\
Traffic Sign   & 0.39                                   & 0.45                                & +15.38 \\
\bottomrule
\end{tabular}}
\end{table}

\begin{table}
\center
  \caption{Impact of Different Segmentation architectures}
  \resizebox{0.5\linewidth}{!}{ 
  \begin{tabular}{ccl}
    \toprule
     &IoU avg   \\
        \midrule
      Proposed  + LeNET & \textbf{0.573}    \\
    \midrule
      Proposed + CENET \cite{9859693}  & 0.567    \\
    \midrule
      Proposed + FIDNET \cite{zhao2021fidnetlidarpointcloud} & 0.558  \\
  \bottomrule
  \end{tabular}}
    \label{tab:lenet}
\end{table}
\textbf{(a) Impact of Resolution on Segmentation performance:} The results in Table~\ref{tab:results_deep} and ~\ref{tab:results_deep_pos} clearly demonstrate how LiDAR resolution significantly influences segmentation performance. Using low-resolution  LiDAR data leads to a noticeable decline in segmentation accuracy. The reduced point density in the low-resolution data results in poorer segmentation. For example, in KITTI dataset the Bicycle class shows a substantial drop of $66\%$, while the Person class experiences a significant decrease of $74\%$. 
\textbf{(b) Independent Training with SOTA Transformer-Based SR \cite{yang2024tulip}}: This scenario utilizes a SOTA LiDAR SR model based on the Swin Transformer architecture, independently trained alongside the segmentation network. The results indicate that while this framework improves segmentation performance for dominant classes, such as cars, it fails to deliver similar benefits for smaller classes like bicycles. The SR model struggles to reconstruct key regions necessary for accurate segmentation of these smaller classes, as the lack of joint optimization between the SR and segmentation networks limits the ability to preserve their  details.
Furthermore, the Transformer-based model contains over 50 million parameters, significantly surpassing the complexity of the segmentation network. This imbalance in complexity highlights the challenges of integrating such models into an end-to-end framework.
\textbf{(c) Proposed End-to-End Framework:} The proposed end-to-end framework achieves superior results with a significant 99\% reduction in parameters compared to the Transformer-based SR model. This is attributed to the lightweight design of the proposed model-based SR framework, making it well-suited for real-time applications. Our model achieves \textbf{23} fps,  outperforming the 6 fps of the method in \cite{yang2024tulip}, as we can see in Table~\ref{tab:results_deep}.  
Also, the framework delivers segmentation performance that is on par with a segmentation network utilizing high-resolution data from a high-cost LiDAR sensor.

\textbf{Impact of the Proposed Context-Aware Loss Function:}  
The results in Table~\ref{tab:context_aware_loss} highlight the impact of the Context-Aware SR Loss on segmentation performance for smaller or underrepresented classes. It is evident that the proposed loss guides the SR model to preserve finer structural details critical for accurate segmentation.

\textbf{Different Segmentation Architectures:} We adopt LeNet as the primary segmentation backbone due to its strong performance. However, our method is architecture-agnostic and can be integrated with other range-view segmentation networks such as CENet~\cite{9859693} and FIDNet~\cite{zhao2021fidnetlidarpointcloud}. As shown in Table~\ref{tab:lenet}, the consistent performance across architectures demonstrates the  generalizability of our approach.

\section{Conclusion}
This work introduced an end-to-end framework for LiDAR SR and semantic segmentation, addressing the limitations of existing methods that often fail to integrate SR and segmentation tasks effectively. Using a novel joint  optimization process and a carefully designed context-aware SR loss function, the proposed framework achieves high segmentation accuracy, particularly for the underrepresented classes, while maintaining computational efficiency. The proposed lightweight LiDAR SR network reduces complexity by $99\%$ compared to state-of-the-art SR models, enabling its seamless integration into an end-to-end system. Experimental results highlight that the framework delivers segmentation performance similar to that of high-resolution  LiDAR data, demonstrating its potential to bridge the gap between affordable low-resolution sensors and high-performance perception systems.

\bibliographystyle{IEEEbib}
\bibliography{icme2025references}

\end{document}